\documentclass[10pt,twocolumn,letterpaper]{article}

\usepackage{cvpr}              

\usepackage{bm}
\usepackage{url}

\newcommand{\SA}{\texttt{SA}}
\def\blue#1{\textcolor{blue}{#1}}

\definecolor{cvprblue}{rgb}{0.21,0.49,0.74}
\usepackage[pagebackref,breaklinks,colorlinks,allcolors=cvprblue]{hyperref}

\def\paperID{15025} 
\def\confName{CVPR}
\def\confYear{2025}

\title{Split Adaptation for Pre-trained Vision Transformers}

\author{Lixu Wang\textsuperscript{1, 2}\footnotemark[1]~, Bingqi Shang\textsuperscript{1}\footnotemark[1]~, Yi Li\textsuperscript{1}, Payal Mohapatra\textsuperscript{1}, Wei Dong\textsuperscript{2}, Xiao Wang\textsuperscript{1}, Qi Zhu\textsuperscript{1}\\
\textsuperscript{1}Northwestern University. 
\textsuperscript{2}Nanyang Technological University.\\
{\tt\small lixu.wang@ntu.edu.sg,\;\{bingqishang2025,\;yili2023.1,\;payalmohapatra2026\}@u.northwestern.edu,}\\
{\tt\small wei\_dong@ntu.edu.sg,\;\{wangxiao,\;qzhu\}@northwestern.edu}}

\begin{document}
\maketitle
\renewcommand{\thefootnote}{\fnsymbol{footnote}}
\footnotetext[1]{Equal contributions.}
\begin{abstract}
Vision Transformers (ViTs), extensively pre-trained on large-scale datasets, have become essential to foundation models, allowing excellent performance on diverse downstream tasks with minimal adaptation. Consequently, there is growing interest in adapting pre-trained ViTs across various fields, including privacy-sensitive domains where clients are often reluctant to share their data. Existing adaptation methods typically require direct data access, rendering them infeasible under these constraints. A straightforward solution may be sending the pre-trained ViT to clients for local adaptation, which poses issues of model intellectual property protection and incurs heavy client computation overhead. To address these issues, we propose a novel split adaptation (\SA{}) method that enables effective downstream adaptation while protecting data and models. \SA{}, inspired by split learning (SL), segments the pre-trained ViT into a frontend and a backend, with only the frontend shared with the client for data representation extraction. But unlike regular SL, \SA{} replaces frontend parameters with low-bit quantized values, preventing direct exposure of the model. \SA{} allows the client to add bi-level noise to the frontend and the extracted data representations, ensuring data protection. Accordingly, \SA{} incorporates data-level and model-level out-of-distribution enhancements to mitigate noise injection's impact on adaptation performance. Our \SA{} focuses on the challenging few-shot adaptation and adopts patch retrieval augmentation for overfitting alleviation. Extensive experiments on multiple datasets validate \SA{}’s superiority over state-of-the-art methods and demonstrate its defense against advanced data reconstruction attacks while preventing model leakage with minimal computation cost on the client side. The source codes can be found at \url{https://github.com/conditionWang/Split_Adaptation}.
\end{abstract}    
\section{Introduction}
\label{sec:intro}
In recent years, pre-trained vision transformers (ViTs)~\cite{vit} have become mainstream foundational models in computer vision, delivering remarkable performance across a range of visual tasks including image classification~\cite{vit}, semantic segmentation~\cite{xie2021segformer, kirillov2023segment}, and multimodal understanding~\cite{xu2023multimodal}. ViTs have been equipped with general semantic knowledge during large-scale pre-training, which can be readily transferred to other domains including sensitive fields such as healthcare~\cite{healthcare}, finance~\cite{financeML}, and manufacturing~\cite{manufacturing}. However, adapting pre-trained ViTs for downstream tasks in these sensitive domains is extremely challenging.

In real-world scenarios, model adaptation often follows a two-party framework, where the pre-trained model is hosted on a server, and the training data resides on a client. In non-sensitive domains, training data can be directly shared with the server, which conducts effective adaptation. Real-world examples include users uploading images to access OpenAI’s CLIP model API~\cite{clip} for data representation or uploading data for fine-tuning the GPT-3.5 model~\cite{gpt}. However, in sensitive domains, data is often tightly linked to personal privacy~\cite{dong2023continual, fang2022shifted}, intellectual property (IP)~\cite{guo2023domain, wang2022non}, and ethical concerns~\cite{gao2024practical}, making any form of public exposure, sharing, or transfer strictly prohibited. In this context, an alternative solution might involve sending the model to the client for local adaptation, as in federated learning~\cite{FL, wang2021addressing, dong2022federated}. However, large pre-trained models are expensive to obtain, involving sophisticated architecture design, extensive pre-training on large-scale high-quality datasets, and lengthy occupying of specialized computing and storage resources. As a result, these models represent valuable IP for their owners, who are unlikely to share or transfer them freely~\cite{wang2022non}. Moreover, given that these pre-trained models consist of billions of parameters, adaptation on the client side would incur substantial computational and storage costs.

\begin{figure}[h]
    \centering
    \includegraphics[width=1.\linewidth]{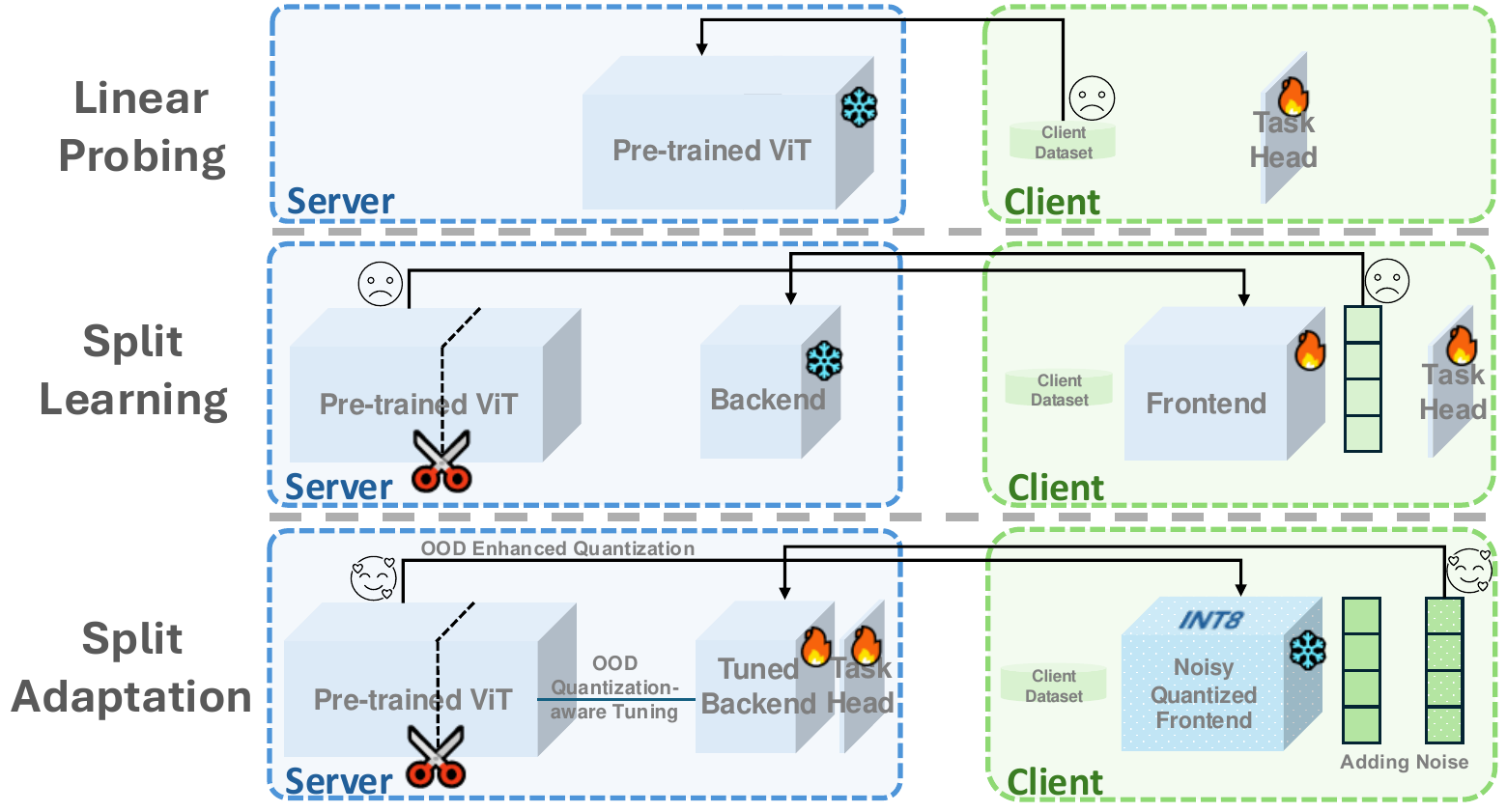}
    \caption{The comparison between different downstream task adaptation approaches. Our \SA{} can achieve effective few-shot task adaptation with minimal computation cost on the client side while protecting both the model and data.}
    \label{fig:teaser}
\end{figure}

To our knowledge, there have been a few preliminary attempts to achieve effective downstream adaptation of pre-trained models while safeguarding data and models. For example, split learning (SL)~\cite{vepakomma2018split, example1} divides a machine learning model into segments managed by different parties, which can partially protect the model from full exposure. Additionally, SL requires only the communication of intermediate representations, avoiding the leakage of raw data. However, SL has critical drawbacks. On one hand, emerging malicious attacks, such as data reconstruction attacks (DRA)~\cite{dra, fsha, gao2023pcat, yin2023ginver, fora}, can infer sensitive information from these intermediate representations. On the other hand, standard SL places the model training on the client side, imposing high computational costs on the client. Beyond SL, a recent method called offsite tuning (OT)~\cite{xiao2023offsite} has been proposed, which involves the server sending an emulator---a simplified version of the pre-trained model---to the client. The client can then fine-tune this emulator using its own data, typically employing parameter-efficient techniques like adapter tuning~\cite{chen2022adaptformer}. After training, the client uploads these adapters back to the server, where they are integrated into the pre-trained model. However, while OT protects data, the emulator’s parameters are fully exposed, allowing an attacker to easily create a high-performance model based on the emulator. Moreover, OT’s adaptation depends heavily on the substantial computational and storage demands for adapter tuning on the client side.

To address the challenge of adapting pre-trained ViTs in sensitive domains, we propose a novel split adaptation (\SA{}) method, which is shown in Figure~\ref{fig:teaser}. Like SL, \SA{} divides the pre-trained ViT into two parts: a frontend and a backend. The server sends the frontend to the client, enabling data representation extraction and model training. However, unlike SL, \SA{} innovatively uses quantization to replace the frontend’s parameters with low-bit values, preventing direct exposure of the original parameters. For additional data protection, the client injects random noise into the received frontend and its extracted data representations. Such injected noise unavoidably hurts the following adaptation, hence \SA{} includes data-level and model-level out-of-distribution enhancements to improve the model generalization when encountering random noise. \SA{} is particularly designed for the challenging few-shot adaptation scenario, where it uses a patch retrieval augmentation technique to generate diverse representations, helping to alleviate overfitting. Extensive experiments across four datasets demonstrate that \SA{} significantly outperforms other state-of-the-art approaches in adaptation performance. Moreover, \SA{} effectively defends against advanced DRAs and prevents model leakage. In particular, \SA{}'s effectiveness only relies on small computation cost on the client side. In summary, our contributions include:
\begin{itemize}[leftmargin=*]
    \item We focus on solving a practical and important problem---adapting pre-trained ViTs for downstream tasks while protecting data and models. Solving this problem enables wider application of pre-trained models in sensitive fields. 
    \item We propose Split Adaptation (\SA{}) that innovatively leverages model splitting and quantization to protect data and models during downstream task adaptation. \SA{} focuses on challenging few-shot scenarios, in which \SA{} uses a patch retrieval augmentation to combat overfitting. 
    \item We conduct extensive experiments across multiple datasets and state-of-the-art baseline approaches, with results validating SA's effectiveness in achieving superior adaptation while ensuring data and model protection.
\end{itemize}

\section{Related Works}
\label{sec:relatedwork}
\subsection{Split Learning}
Split learning (SL)~\cite{vepakomma2018split} is a collaborative learning approach widely used in the multi-party computation framework. SL operates by dividing the machine learning (ML) model into segments, with each segment managed by a different party. This approach allows the computational resources of different parties to be utilized more efficiently. More importantly, SL enhances privacy by transmitting only intermediate data representations rather than raw data. Many practical examples have demonstrated the effectiveness of SL in training and adapting ML models across various domains including healthcare~\cite{vepakomma2018split, example1}, financial~\cite{finance1, example1}, and edge computing~\cite{IoT, IoT1}. Despite these successes, more and more adversarial attacks illustrate the vulnerability of privacy leakage in SL, with the severest one -- data reconstruction attack (DRA)~\cite{dra}. DRA aims to recover the raw data from its corresponding model representations. In the context of SL, a variety of DRAs have been witnessed, among them, some are based on optimization to imitate the ground-truth representations (e.g., FSHA~\cite{fsha} and PCAT~\cite{gao2023pcat}), while others leverage generative models to recover the raw data by exploiting the intrinsic semantics (e.g., Ginver~\cite{yin2023ginver} and FORA~\cite{fora}). Therefore, new SL approaches are needed to help better protect data privacy.

\subsection{Vision Transformer Adaptation}
The vision transformers (ViT)~\cite{vit} are usually pre-trained with large-scale data to learn general semantics and knowledge. Representative pre-training approaches include supervised and self-supervised learning (SSL), e.g., SAM~\cite{kirillov2023segment} is trained on a large amount of pixel-level annotated datasets; a variety of SSL methods, like contrastive learning~\cite{simclr, moco, dino, clip} and auxiliary task training~\cite{puzzle, gan_ssl, mae}. Owing to such extensive pre-training, adapting pre-trained ViT for distinct downstream tasks has become a mainstream paradigm. The most regular method is to fine-tune the entire model, which consumes substantial computational and memory resources. Another widely used approach is linear probing~\cite{linear_probing}, which only trains a task module while keeping the ViT frozen, but the prominent shortcoming is the inferior performance. Recently, more parameter-efficient fine-tuning approaches have been proposed, like adapter tuning~\cite{chen2022adaptformer}, prompt tuning~\cite{vpt}, and prefix tuning~\cite{prefix}. However, these approaches still require direct access to the task data. In contrast to the above approaches, a new method called offsite tuning~\cite{xiao2023offsite} can work effectively without direct data access. However, extensive training happens on the data owner side, and there is a high risk of model disclosure, making it impractical in real-world scenarios.

\section{Methodology}

\subsection{Preliminaries}

\subsubsection{Vision Transformer}
\label{sec:vit}
The vision transformer (ViT)~\cite{vit} is built upon sequential attention encoder layers. Specifically, each encoder layer consists of a multi-head self-attention (MSA) block and a Multi-Layer Perceptron (MLP) block. Layernorm (LN) is applied before every block and residual connections are placed after every block. The MLP contains two linear layers with an intermediate GELU activation. To handle 2D images, ViT reshapes the image $\mathbf{I} \!\in\! \mathbb{R}^{H \times W \times C}$ into $N$ flattened patches $I^p \!\in\! \mathbb{R}^{P^2 \cdot C}$, where $H$, $W$, and $C$ are the image height, width, and channels, respectively, and $P$ is the pixel resolution of each patch. For the first layer of ViT, to unify the vector dimensions, a trainable linear projection layer with neuron weights $[\mathbf{W}_1^E, \cdots, \mathbf{W}_N^E]$ is used to map each vectorized patch into dimension $d$. In this case, the input to the first transformer layer is:
\begin{equation}
\begin{aligned}
    &\mathbf{X}_1 = [\mathbf{x}^\mathrm{CLS};I_1^p\mathbf{W}_1^E; \cdots; I_N^p\mathbf{W}_N^E] + \mathbf{E}^\mathrm{POS},\\
    &\text{where}\, \mathbf{W}^E \in \mathbb{R}^{(P^2\cdot C)\times d}, \mathbf{E}^\mathrm{POS} \in \mathbb{R}^{(N+1) \times d},
\end{aligned}
\end{equation}
where $\mathbf{x}^\mathrm{CLS} \!\in\! \mathbb{R}^d$ is a classification token, and $\mathbf{E}^\mathrm{POS}$ is the position encoding. Then for each of the following layers, the input sequence $\mathbf{X}_l$ is fed into each head of the MSA module to obtain three linearly projected matrices: Query $\mathbf{Q}_l \!=\! \mathbf{X}_l \mathbf{W}_l^Q$, Key $\mathbf{K}_l \!=\! \mathbf{X}_l \mathbf{W}_l^K$, and Value $\mathbf{V}_l \!=\! \mathbf{X}_l \mathbf{W}_l^V$. Then the Softmax function is applied to normalize the attention score $\mathbf{A}_l \!=\! \mathbf{Q}_l \mathbf{K}_l^\top$ and the output $\mathbf{Z}_l$ is:
\begin{equation}
    \mathbf{Z}_l = \mathrm{MSA}(\mathbf{X}_l) = \mathrm{Softmax}\left(\frac{\mathbf{A}_l}{\sqrt{d}}\right)\mathbf{V}_l.
\end{equation}
Suppose the two linear layers in MLP are parameterized by $\mathbf{W}_{l, 1}, \bm{b}_{l, 1}$ and $\mathbf{W}_{l, 2}, \bm{b}_{l, 2}$, the output is then computed as:
\begin{equation}
    \mathrm{MLP}(\mathbf{Z}_l) = \mathrm{GELU}(\mathbf{Z}_l \mathbf{W}_{l, 1} + \bm{b}_{l, 1}) \mathbf{W}_{l, 2} + \bm{b}_{l, 2}.
\end{equation}
Finally, both the MSA and MLP modules are linked via residual connections, thus we have the final output of each transformer layer: $\mathbf{X}_{l+1} \!=\! \mathrm{LN}(\mathbf{Z}_l + \mathrm{MLP}(\mathbf{Z}_l)) \!\in\! \mathbb{R}^{(N+1)\times d}$, where $\mathbf{Z}_l \!=\! \mathrm{LN}(\mathbf{X}_l + \mathrm{MSA}(\mathbf{X}_l)) \!\in\! \mathbb{R}^{(N+1)\times d}$.

\subsubsection{Problem Formulation}
When adapting pre-trained ViT for certain downstream tasks, we consider a representative client-server framework in which the client queries the server for the adaptation request. The client owns the task adaptation dataset $\mathcal{D}^\mathrm{C} \!=\! \{(\bm{x}_i, y_i)\}_{i=1}^{N^\mathrm{C}}$ ($N^\mathrm{C}$ is the sample number, in practice, we suppose $N^\mathrm{C}$ is small even few-shot) with each sample drawn from a marginal input distribution $\bm{x}_i \!\sim\! \mathcal{P}_\mathcal{X}^\mathrm{C}$ and labeled with a marginal task distribution $\mathcal{P}_\mathcal{Y}^\mathrm{C}$. 

On the server side, the pre-trained ViT $F_\Theta$ parameterized on $\Theta$ is assumed to consist of $L^F$ attention layers, and each layer is parameterized on $\theta$, i.e., $F\!:=\![f_{\theta_1}, \cdots, f_{\theta_{L^F}}]$. Besides, we assume the server owns another dataset $\mathcal{D}^\mathrm{S} \!=\! \{(\bm{x}_i, y_i)\}_{i=1}^{N^\mathrm{S}}$ collected from public domains with the marginal input and task distributions, $\mathcal{P}_\mathcal{X}^\mathrm{S}$ and $\mathcal{P}_\mathcal{Y}^\mathrm{S}$, respectively. In real-world scenarios, the datasets owned by the client and the server are distinct but relevant, i.e., $\mathcal{P}_\mathcal{X}^\mathrm{C} \!\neq\! \mathcal{P}_\mathcal{X}^\mathrm{S}, \mathcal{P}_\mathcal{Y}^\mathrm{C} \!\neq\! \mathcal{P}_\mathcal{Y}^\mathrm{S}$ but $I(\mathcal{X}^\mathrm{C}; \mathcal{X}^\mathrm{S}) \!>\! 0, I(\mathcal{Y}^\mathrm{C}; \mathcal{Y}^\mathrm{S}) \!>\! 0$ where $I(\cdot; \cdot)$ calculates the mutual information.

\smallskip \noindent \textbf{Objective of Task Performance.} 
Pre-trained ViT adaptation aims to obtain a downstream task module $g$ parameterized on $\gamma$ that can achieve as good performance as possible on the client data distribution when combined with $F_\Theta$. Certainly, such adaptation also allows $F_\Theta$ tuning if it can enable performance improvement on the downstream task. In this case, the objective is formulated as,
\begin{equation}
    \Theta, \gamma = \arg \min_{\Theta, \gamma} \mathbb{E}_{\bm{x} \sim \mathcal{P}_\mathcal{X}^\mathrm{C}, y \sim \mathcal{P}_\mathcal{Y}^\mathrm{C}} \mathcal{L}(\bm{x}, y; \Theta \circ \gamma),
\label{eq:objective}
\end{equation}
where $\mathcal{L}(\cdot;\cdot)$ is a loss function that measures the model performance regarding certain data distribution. 

\begin{figure*}[ht]
    \centering
    \includegraphics[width=.95\linewidth]{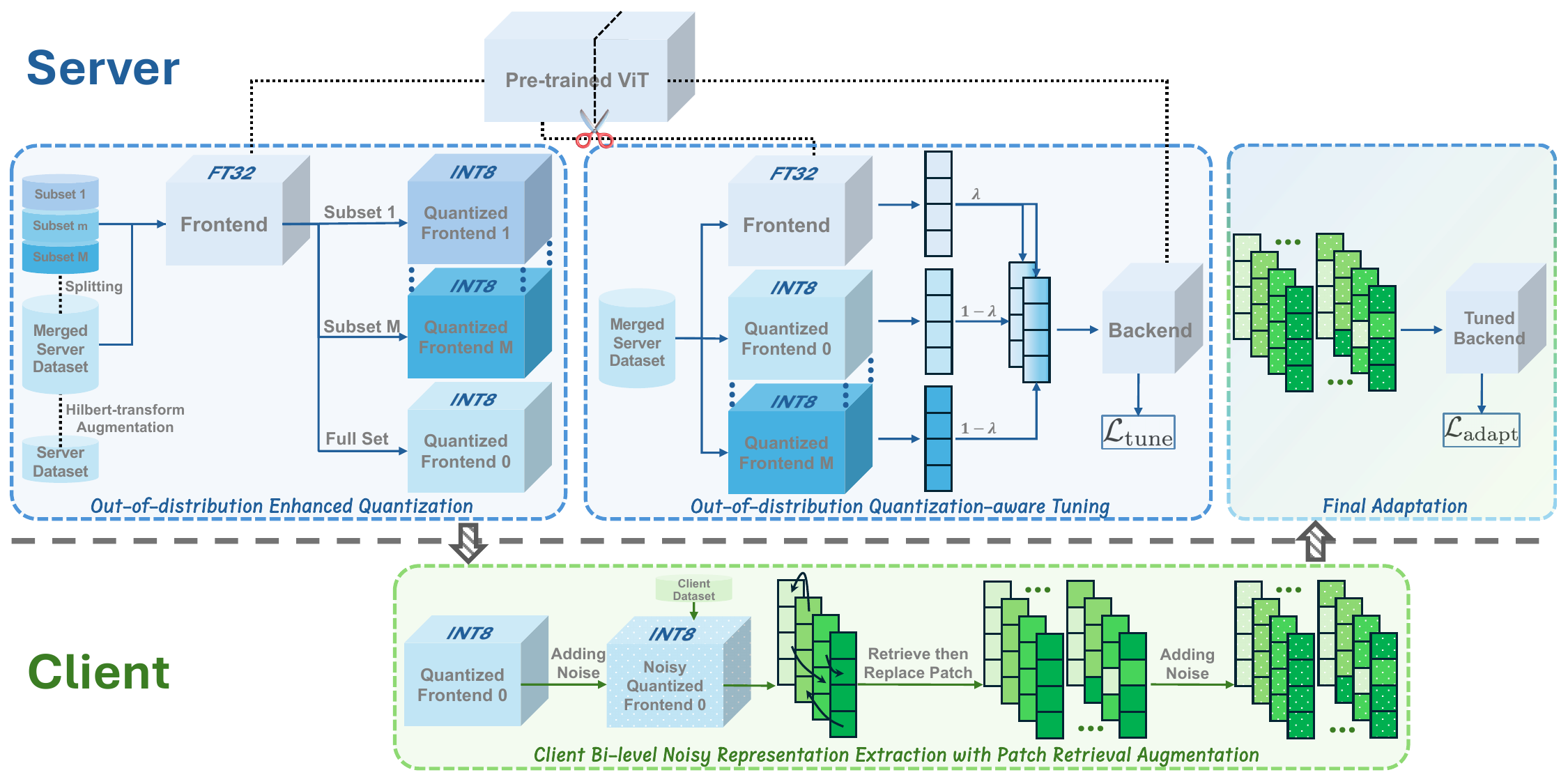}
    \vspace{-10pt}
    \caption{Overview of Split Adaptation (\SA{}) for pre-trained ViT adaptation. \SA{} divides the pre-trained ViT into a frontend and a backend. After applying Out-of-distribution Enhanced Quantization to the frontend, its quantized version is sent to the client. To mitigate the impact of quantization, \SA{} adopts Out-of-distribution Quantization-aware Tuning to enhance backend's generalization. As for the client, it injects random noise to the received frontend and retrieves then replaces randomly selected patches to augment more client data representations, which are sent to the server for the final adaptation after being added with noise again.}
    \label{fig:fig1}
\end{figure*}

\smallskip \noindent \textbf{Objective of Privacy Protection.}
In addition to the above standard objective, the intrinsic privacy concerns of the server-client framework should be addressed, i.e., the pre-trained model and client data should be protected during adaptation when supposing the server and the client may be semi-honest. On the one hand, the adaptation should prohibit the client from stealing surrogate models with good performance or obtaining them at a small cost, on the other hand, any client data reconstruction on the server side should be avoided. 

\subsection{Overview of The Proposed Split Adaptation}
To achieve effective adaptation while protecting the pre-trained ViT and client data, we propose a method called Split Adaptation (\SA{}), shown in Figure~\ref{fig:fig1}. Our \SA{} starts with dividing the pre-trained ViT $F_\Theta$ into two parts: a frontend $F^\mathrm{F}$ and a backend $F^\mathrm{B}$. The original $F^\mathrm{F}$ is not sent to the client, instead, we leverage model quantization to send a quantized version $\widehat{F}^\mathrm{F}$ to the client. To mitigate the influence of frontend quantization, \SA{} includes out-of-distribution (OOD) augmentations from both data and model levels to enhance the cross-distribution generalization of the frontend and backend. As for the client side, it adds random noise to the received $\widehat{F}^\mathrm{F}$ first and adopts a patch-level retrieve-then-replace strategy to generate more client data representations that help alleviate overfitting in few-shot adaptation. Before being uploaded to the server for the final adaptation, these augmented representations are protected by adding noise again.

\subsection{Model Splitting on The Server Side}
At the beginning of \SA{}, the pre-trained ViT is divided into a frontend $F^\mathrm{F}\!:=\![f_{\theta_1}, \cdots, f_{\theta_K}]$ and a backend $F^\mathrm{B}\!:=\![f_{\theta_{K\!+\!1}}, \cdots, f_{\theta_{L^F}}]$ at the $K$-th layer ($K \!>\! L^F / 2$). Then the frontend is distributed to the client for representation extraction. Although we empirically found that the frontend cannot perform well (Section~\ref{sec:exp_privacy_protection}), we still hope to protect it from disclosing the original model parameters. Quantization~\cite{symmetric_quant, quantization1}, as a representative model compression approach, can significantly reduce the computational and memory cost without sacrificing model performance. As quantization works by representing model weights and activations with low-precision values, it can potentially protect the model. However, standard quantization needs a certain amount of data drawn from the target task, which is inaccessible for the server in our problem. In this case, the left choice is to leverage the server dataset $\mathcal{D}^\mathrm{S}$ to quantize the model, but a dedicated design for enhancing the quantized model's cross-distribution generalization is needed. Therefore, we propose an augmentation technique to generate OOD data in terms of $\mathcal{D}^\mathrm{S}$ for quantization.
\begin{figure}[t]
    \centering
    \includegraphics[width=1.\linewidth]{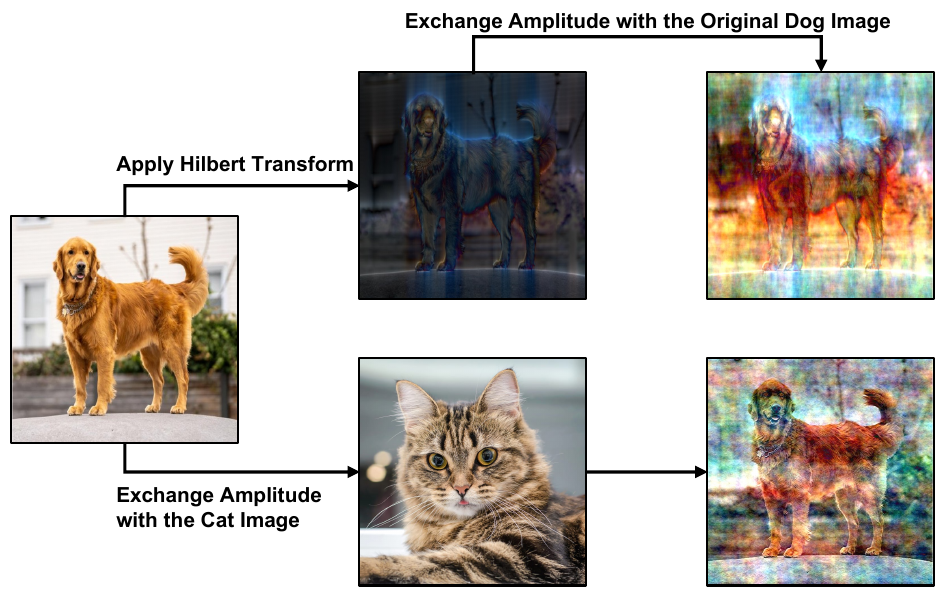}
    \vspace{-25pt}
    \caption{Visualization of our Hilbert Transform data augmentation. Compared to amplitude-exchange augmentation~\cite{wang2022domain} (bottom row), our augmentation method (top row) generates data that diverges further from the original in appearance while preserving the original semantics (dog category).}
    \label{fig:ht_aug}
    \vspace{-15pt}
\end{figure}

\subsubsection{Out-of-distribution Enhanced Quantization}
\label{sec:ood_quantization}
In practice, we have to consider extreme cases in which the client dataset differs significantly from the server dataset and they may only share general semantics. Hence our augmentation technique should generate OOD data with a large discrepancy to $\mathcal{D}^\mathrm{S}$. Inspired by signal processing, we notice that there is a technique called Hilbert Transform (HT), which only captures the most sharp edge in images. Intuitively, the sharp edges contain the most general semantics, making HT suitable for our problem. Applying HT to a signal in standard signal processing results in a phase shift of $\pi/2$ in the frequency domain~\cite{mohapatra2024phase}. In the visual domain, we can implement HT by transforming an image $\mathbf{I}$ to the frequency domain with the spectrum index $\xi$ via Fourier transform, then the negative $\xi$ needs to multiply with the imaginary unit $i$, while the positive $\xi$ needs to multiple with $-i$, i.e., 
\begin{equation}
    \mathrm{HT}(\mathbf{I}) = \mathcal{F}^{-1}\{-i \cdot \mathrm{sgn}(\xi)\mathcal{F}(\mathbf{I})\},
\end{equation}
where $\mathcal{F}(\cdot)$ and $\mathcal{F}^{-1}(\cdot)$ are Fourier transform and inverse, respectively, and $\mathrm{sgn}(\cdot)$ is a sign function. However, pure HT only provides new images with black backgrounds. To diversify the HT-generated image, we replace its amplitude in the frequency domain with that of the original image. In this way, the generated image is much better than the standard amplitude-exchange augmentation~\cite{wang2022domain, liudeja} (please refer to Figure~\ref{fig:ht_aug} for a visual comparison).

To achieve effective quantization, we adopt the most popular uniform symmetric quantization~\cite{symmetric_quant} to implement \SA{}, in which a floating-point value $w$ is projected into a $k$-bit integer value $w_\mathrm{q}$ with a scaling factor $\Delta$:
\begin{equation}
    \widehat{w} = \Psi_k(w, \Delta) = \mathrm{clamp}(\mathrm{round}(\frac{w}{\Delta}), -2^{k-1}, 2^{k-1}-1),
\label{eq:value_quantization}
\end{equation}
where $\mathrm{round}(\cdot)$ turns a value into an integer and $\mathrm{clamp}(\cdot, min, max)$ constrains the output in the range from $min$ to $max$. Intuitively, the scaling factor $\Delta$ directly impacts the model performance after quantization. Previous studies~\cite{quant_metric1, quantization1} have tried a variety of measurements to quantify the model output changes and accordingly determine the optimal scaling factors. Among these studies, we follow Yuan et al.~\cite{yuan2022ptq4vit} to choose a Hessian-guided metric for scaling factor determination. Suppose the pre-trained ViT $F_\Theta$ is equipped with a task module $g_{\gamma^\prime}^\prime$ that is tuned on the server dataset $\mathcal{D}^\mathrm{S}$, the model performance can be measured by a loss $\mathbb{E}_{\bm{x} \sim \mathcal{P}_\mathcal{X}^\mathrm{S}, y \sim \mathcal{P}_\mathcal{Y}^\mathrm{S}} \mathcal{L}(\bm{x}, y; \Theta \circ \gamma^\prime)$ (we omit $\bm{x}$ and $y$ to simplify the used symbols in the following descriptions). Then if we regard that the quantization brings a small perturbation $\epsilon$ to the model parameters, i.e., $\widehat{\Theta} = \Theta + \epsilon$, we can analyze the influence of quantization on the task loss using Taylor series expansion:
\begin{equation}
    \mathbb{E} [\mathcal{L}(\widehat{\Theta} \circ \gamma^\prime)] - \mathbb{E} [\mathcal{L}(\Theta \circ \gamma^\prime)] \approx \epsilon^\top \nabla_\Theta + \frac{1}{2}\epsilon^\top\mathbf{H}_\Theta \epsilon,
\label{eq:hessian_metric}
\end{equation}
where $\nabla_\Theta$ is the gradients and $\mathbf{H}_\Theta$ is the Hessian matrix. Then the optimal scaling factors are the ones corresponding to minimal influence caused by quantization, i.e., $\Delta = \arg \min_\Delta (\mathbb{E} [\mathcal{L}(\widehat{\Theta} \circ \gamma^\prime)] - \mathbb{E} [\mathcal{L}(\Theta \circ \gamma^\prime)])$. Certainly, determining and adopting a general scaling factor for the entire model is too coarse to perform well, thus we apply a layer-wise reconstruction method~\cite{quant_reconstruct} to achieve fine-grained quantization, then the searching optimization of Eq.~\eqref{eq:hessian_metric} can be approximated by:
\begin{equation}
    \arg \min_{\Delta_{l-1}} \mathbb{E}\left[(\widehat{\mathbf{X}}_{l} - \mathbf{X}_{l})^\top \mathrm{diag}\left(\frac{\partial \mathcal{L}}{\partial \mathbf{X}_{l}} \odot \frac{\partial \mathcal{L}}{\partial \mathbf{X}_{l}}\right) (\widehat{\mathbf{X}}_{l} - \mathbf{X}_{l})\right],
\label{eq:layer_quantization}
\end{equation}
where $\mathbf{X}_{l}$ and $\widehat{\mathbf{X}}_{l}$ are the outputs of the $(l\!-\!1)$-th layer before and after quantization, respectively, and $\odot$ represents the Hadamard product of a matrix.

With the augmented neighborhood data, the frontend can be quantized with improved cross-distribution generalization. Specifically, we first conduct HT-based data augmentation on the server dataset $\mathcal{D}^\mathrm{S}$ to obtain $\mathrm{HT}(\mathcal{D}^\mathrm{S})$. Then the optimal scaling factor for each layer in the frontend $F^\mathrm{F}$ is empirically determined by optimizing Eq.~\eqref{eq:layer_quantization} on the merged dataset $\mathcal{D}^\mathrm{S}_\cup = \mathcal{D}^\mathrm{S} \cup \mathrm{HT}(\mathcal{D}^\mathrm{S})$ and we can obtain a quantized frontend $\widehat{F}^\mathrm{F}$ in the end.

\subsubsection{Out-of-distribution Quantization-aware Tuning}
\label{sec:backend_tuning}
To achieve better adaptation performance, the backend also needs to resist the unavoidable parameter difference between the original and quantized model as well as the noise injected by the client that we will introduce in Section~\ref{sec:client_side}. As a result, we propose an OOD quantization-aware tuning technique to enhance the backend's resistance when encountering OOD data and frontends.

To simulate potential OOD frontends, we first divide the merged server dataset $\mathcal{D}^\mathrm{S}_\cup$ into $M$ subsets and denote these subsets as $\{\mathcal{D}^\mathrm{S}_{\cup_m}\}_{m=1}^M$. Next, each subset $\mathcal{D}^\mathrm{S}_{\cup_m}$ is used to quantize the original frontend $F^\mathrm{F}$ to obtain a quantized version $\widehat{F}^\mathrm{F}_m$. Then let us use a single quantized frontend $\widehat{F}^\mathrm{F}_m$ to introduce how to tune the backend. To begin with, since the backend is represented and the tuning is performed using floating-point values, the data representations extracted by the quantized frontend should be transformed back to floating-point values through the dequantization:
\begin{equation}
    \widetilde{w} = \Psi^{-1}(\widehat{w}, \Delta) = \widehat{w} \cdot \Delta.
\label{eq:dequantization}
\end{equation}
In this way, we can obtain the floating-point representation $\widetilde{\mathbf{X}}_K$ by dequantizing the $K$-th layer's output with the scaling factor $\Delta_K$. Note that $\widetilde{\mathbf{X}}_K$ is different from the original representation $\mathbf{X}_K$ extracted by the original frontend $F^\mathrm{F}$ due to the range constraints of $\mathrm{round}$ and $\mathrm{clamp}$ functions in Eq.~\eqref{eq:value_quantization}. Our following design aims to tune the backend to resist such differences. The most straightforward method is to forward the floating-point representation $\widetilde{\mathbf{X}}_K$ to the backend and conduct tuning using task loss. However, such tuning will make the backend over-fitted to the server dataset $\mathcal{D}^\mathrm{S}$. We solve this by a representation-mixup strategy. Specifically, for a particular data sample $\bm{x}$, we feed it to a quantized frontend $\widehat{F}^\mathrm{F}_m$ and the original frontend $F^\mathrm{F}$ to extract representations $\widehat{\mathbf{X}}_{K\!, m}$, $\mathbf{X}_{K\!, m}$. After dequantizing $\widehat{\mathbf{X}}_{K\!, m}$ into the floating-point form $\widetilde{\mathbf{X}}_{K\!, m}$, we obtain a mixup representation:
\begin{equation}
    \mathbf{X}_{K\!, m}^* = \lambda \widetilde{\mathbf{X}}_{K\!, m} + (1 - \lambda)\mathbf{X}_{K\!, m},
\end{equation}
where $\lambda$ is drawn from a widely-used beta distribution $\mathrm{Beta}(\lambda;0.75, 0.75)$. Then considering all available quantized frontend, we can tune the backend with the loss:
\begin{equation}
    \mathcal{L}_\mathrm{tune} = \sum_{\mathcal{D}^\mathrm{S}_\cup} \sum_{m=0}^M \mathcal{L}(\mathbf{X}^*_{K\!, m}, y; \Theta^\mathrm{B} \circ \gamma^\prime).
\end{equation}
Note that $\mathbf{X}_{K\!, 0}^*$ originates from feeding $\bm{x}$ to the frontend $\widehat{F}^\mathrm{F}$ quantized on the full set of $\mathcal{D}_\cup^\mathrm{S}$.

\subsection{Client Bi-level Noisy Representation Extraction}
\label{sec:client_side}
In our \SA{} model, the quantized frontend $\widehat{F}^\mathrm{F}$ will be sent to the client after the OOD enhanced quantization (Section~\ref{sec:ood_quantization}), and then the client can forward its data to $\widehat{F}^\mathrm{F}$ to extract representations and send them to the server. However, there is a risk of data reconstruction when sharing representations, thus we propose a bi-level noisy representation extraction mechanism to defend against such attacks.

Intuitively speaking, the key to data reconstruction defense is to use the model unknown by the server to extract the representations. Therefore, we allow the client to inject a certain degree of random noise into the received frontend $\widehat{F}^\mathrm{F}$. For each layer in the frontend, we inject a multiplicative noise $n_l^\times$, and an additive noise $n_l^+$, which are sampled from two Gaussian distributions, i.e., $n_l^\times \!\sim\! \mathcal{P}_{\mathcal{N}(1, \alpha \theta_l)}$ and $n_l^+ \!\sim\! \mathcal{P}_{\mathcal{N}(0, \alpha \theta_l)}$ ($\alpha\!=\!0.01$ controls the noise degree), to the model parameters:
\begin{equation}
    \theta_l^\prime = n_l^\times \cdot \theta_l + n_l^+.
\end{equation}
Next, the client feeds its dataset $\mathcal{D}^\mathrm{C}$ to the noisy frontend $\widehat{F}^{\mathrm{F}^\prime}$ to extract representations $\{\widehat{\mathbf{X}}_{K\!,i}^{\mathrm{C}}\}_{i=1}^{N^\mathrm{C}}$. However, it is still insecure to share these representations. We found that our model-level noise can amplify the defense effect of the regular noise obfuscation~\cite{he2020attacking} at the representation level, which allows to add a smaller degree of Laplace noise ($\mathcal{P}_{\mathrm{Laplace}(0, 0.8)}$) to representations for better protection.

\subsection{Patch-retrieval-augmented Model Adaptation}
\label{sec:patch_aug}
In few-shot adaptation scenarios, like the backend tuning on the server (Section~\ref{sec:backend_tuning}), the adaptation also faces overfitting challenges, but the overfitting here is due to the limited client data available. To tackle such challenges, we design a patch-level retrieval augmentation approach to generate more client data representations.

\begin{table*}[htbp]
\centering
\caption{Performance comparison between our \SA{} and other baseline approaches in 3-shot, 5-shot, and 10-shot adaptation scenarios over 3 datasets. The \SA{} can substantially exceed other methods. We bold and blue \blue{\textbf{the best}}, and bold \textbf{the second best}.}
\vspace{-8pt}
\hspace{-8pt}
\resizebox{.95\textwidth}{!}{
\setlength{\tabcolsep}{1.2mm}{
\centering
\begin{tabular}{l|ccc|ccc|ccc}
\toprule
Few-shot Setup & \multicolumn{3}{c|}{3-shot} & \multicolumn{3}{c|}{5-shot} & \multicolumn{3}{c}{10-shot}\\ \midrule
Method & CIFAR-100 & Places365 & D.Net-Cl & CIFAR-100 & Places365 & D.Net-Cl & CIFAR-100 & Places365 & D.Net-Cl \\ \midrule
Linear Probing & $\bm{74.05}_{\pm 1.21}$ & $26.18_{\pm 1.00}$ &$\bm{57.26}_{\pm 1.11}$  &$78.79_{\pm0.65}$  &$\bm{34.25}_{\pm1.45}$  &$63.84_{\pm0.66}$  &$83.28_{\pm0.33}$  &$32.33_{\pm7.82}$  &$\bm{69.12}_{\pm0.75}$    \\
Fine Tuning & $40.32_{\pm 5.48}$ & $21.64_{\pm 0.78}$ & $36.95_{\pm 6.66}$ &$64.29_{\pm7.69}$  &$31.07_{\pm1.04}$  &$56.50_{\pm1.87}$  &$81.82_{\pm1.85}$  &\blue{$\bm{39.46}_{\pm0.59}$}  &$70.67_{\pm0.72}$    \\
LN-TUNE &$13.18_{\pm 2.17}$  & $1.59_{\pm 0.53}$  & $6.97_{\pm 1.96}$ &$34.80_{\pm5.37}$  &$3.96_{\pm0.95}$  &$13.43_{\pm2.12}$  &$47.97_{\pm6.25}$  &$7.71_{\pm1.85}$  &$24.53_{\pm2.05}$    \\
Split Learning & $\bm{74.05}_{\pm 1.21}$ & $\bm{26.84}_{\pm 1.03}$ & \blue{$\bm{58.14}_{\pm 0.63}$} &$\bm{79.63}_{\pm0.53}$  &$30.49_{\pm0.36}$  &$\bm{63.95}_{\pm1.18}$  &$\bm{83.66}_{\pm0.27}$  &$35.78_{\pm0.30}$  &$65.20_{\pm0.47}$    \\
Offsite Tuning & $42.37_{\pm 2.98}$ & $24.98_{\pm 0.55}$ & $40.56_{\pm 4.00}$  &$64.83_{\pm6.17}$  &$30.45_{\pm0.80}$  &$56.43_{\pm0.45}$  &$80.11_{\pm1.07}$  &$36.22_{\pm1.33}$  &$68.43_{\pm0.64}$    \\ \midrule
\SA{} (ours) & \blue{$\bm{76.24}_{\pm 0.29}$}  & \blue{$\bm{30.92}_{\pm 0.89}$} & $56.26_{\pm 0.79}$  &\blue{$\bm{81.98}_{\pm0.49}$}  &\blue{$\bm{35.31}_{\pm0.01}$}  &\blue{$\bm{65.03}_{\pm0.59}$}  &\blue{$\bm{85.45}_{\pm0.40}$}  &$\bm{39.26}_{\pm0.10}$  &\blue{$\bm{71.13}_{\pm0.76}$}    \\ \bottomrule
\end{tabular}}}
\label{tab:5-shot}
\vspace{-5pt}
\end{table*}

Our patch-level retrieval augmentation is inspired by the intuition that different images may contain similar patches. As introduced in Section~\ref{sec:vit}, the data representations of each layer in a ViT hold the dimension $(N\!+\!1)\!\times\! d$, which can be denoted as $\widehat{\mathbf{X}}^{\mathrm{C}}_{K\!, i}\!=\![\mathbf{x}^\mathrm{CLS}_i; \mathbf{x}_{i, 1}, \cdots, \mathbf{x}_{i, N}]$ (we omit some symbols for simplification). Then we can form $N$ retrieval sets from all client data representations (before adding Laplace noise) and each corresponds to a patch, i.e., $\{\mathcal{S}_j\}_{j=1}^N$ where $\mathcal{S}_j \!=\! \{\mathbf{x}_{i, j}\}_{i=1}^{N^\mathrm{C}}$. With these sets, for each augmentation of $\widehat{\mathbf{X}}^{\mathrm{C}}_{K\!, i}$, we randomly select $N^\mathrm{P} \!=\! 50$ patches (please see the sensitivity analysis in Supplementary Materials) $\{\mathbf{x}_{i, \mathrm{P}_j}\}_{j=1}^{N^\mathrm{P}}$ (subscript $\mathrm{P}_j$ meets $1 \!\leq\! \mathrm{P}_j \!\leq\! N$) and delete them from their corresponding retrieval sets. Next, each $\mathbf{x}_{i, \mathrm{P}_j}$ retrieves the most similar patch $\Omega(\mathbf{x}_{i, \mathrm{P}_j})$ from its corresponding retrieval set:
\begin{equation}
    \Omega(\mathbf{x}_{i, \mathrm{P}_j})= \arg \max_{\mathrm{x} \in \mathcal{S}^\prime_{\mathrm{P}_j}} \frac{\mathbf{x}_{i, \mathrm{P}_j} \cdot \mathbf{x}}{|\mathbf{x}_{i, \mathrm{P}_j}||\mathbf{x}|},
\end{equation}
where $\mathcal{S}^\prime_{\mathrm{P}_j} \!=\! \mathcal{S}_{\mathrm{P}_j} \!\setminus\! \mathbf{x}_{i, \mathrm{P}_j}$. After the retrieval operations for all selected patches $\{\mathbf{x}_{i, \mathrm{P}_j}\}_{j=1}^{N^\mathrm{P}}$, we replace them with the retrieved patches $\{\Omega(\mathbf{x}_{i, \mathrm{P}_j})\}_{j=1}^{N^\mathrm{P}}$ in the original representation $\widehat{\mathbf{X}}^{\mathrm{C}}_{K\!, i}$ to form a new one $\widehat{\mathbf{X}}^{^\prime\mathrm{C}}_{K\!, i}$. Note that the selected patches are determined randomly, resulting in newly generated representations that differ across augmentation runs. To generate sufficient augmented representations, we run the augmentation $N^\mathrm{Aug}$ times for all client data representations and produce an augmented representation set, i.e., $\widehat{\mathcal{X}}^\mathrm{C}_K \!=\! \{\ \{\widehat{\mathbf{X}}_{K\!, i}^{\mathrm{C}_j}\}_{j=0}^{N^\mathrm{Aug}}\}_{i=1}^{N^\mathrm{C}}$ (we denote the original representation as $\widehat{\mathbf{X}}_{K\!, i}^{\mathrm{C}_0}$ and the augmented representations as $\widehat{\mathbf{X}}_{K\!, i}^{\mathrm{C}_{1 \sim N^\mathrm{Aug}}}$). The above patch retrieval augmentation is run by the client, and it adds Laplace noise (Section~\ref{sec:client_side}) to all representations in $\widehat{\mathcal{X}}^\mathrm{C}_K$ before uploading to the server.

After receiving the representations, the server dequantizes them into floating-point form with the $K$-th layer's scaling factor $\Delta_K$ by Eq.~\eqref{eq:dequantization} and launches the adaptation. To further alleviate the overfitting due to limited client data, different from Split Learning, we only allow high-level feature learning in the backend to prevent low-level features from becoming biased. Then the server randomly initializes a task module $g_\gamma$ and tunes it with the backend $F^\mathrm{B}$ (that has been tuned in Section~\ref{sec:backend_tuning}) by following the loss:
\begin{equation}
    \mathcal{L}_\mathrm{adapt} = \sum_{i=1}^{N^\mathrm{C}} \sum_{j=0}^{N^\mathrm{Aug}} \mathcal{L}(\widehat{\mathbf{X}}^{\mathrm{C}_j}_{K\!, i}, y; \Theta^\mathrm{B} \circ \gamma).
\end{equation}
To protect the client’s labels, the server sends the task module output to the client, allowing the task loss $\mathcal{L}_\mathrm{adapt}$ to be computed on the client side and then sent back to the server.

\section{Experiments}
\subsection{Experimental Settings}
\noindent \textbf{Datasets.} \textit{ImageNet}~\cite{deng2009imagenet} is the most widely used image dataset collected from public domains, thus we use its test set as the server dataset. For the client datasets, we use \textit{CIFAR-100}~\cite{cifar}, \textit{Places365}~\cite{places365}, and \textit{DomainNet}~\cite{domainnet}. Places365 is a scene recognition dataset that contains 365 classes. DomainNet (D.Net) is a challenging multi-domain dataset, which includes six domains. We select the smallest domain---Clipart (Cl) for the few-shot adaptation scenarios.

\smallskip \noindent \textbf{Implementation Details.} We conduct experiments on three pre-trained model architectures: ViT-Large~\cite{vit}, DeiT-Base~\cite{deit}, and Swin-L~\cite{liu2021swin} (DeiT-Base and Swin-L's experiments are reported in Supplementary Materials). For the Places365 dataset, the target task module is composed of two linear layers, whereas for other datasets, it consists of a single linear layer. We divide 2/3 backbone model as the fontend while the rest 1/3 is the backend. The frontend and its inner activations are quantized to 8-bit. Each quantization operation uses 32 data samples randomly sampled from the server dataset or subsets, and the batch size is 4. The Adam optimizer with a learning rate of $10^{-5}$ is adopted. The number of epochs for OOD quantization-aware tuning is set to 1, while the final adaptation is set to 100 epochs. All experiments are run repeatedly 3 times with three seeds, and we report the mean performance and standard deviation.

\smallskip \noindent \textbf{Baseline Approaches.} \SA{} is compared with state-of-the-art downstream task adaptation methods including Linear Probing~\cite{linear_probing}, Fine Tuning, Split Learning~\cite{vepakomma2018split}, Offsite Tuning~\cite{xiao2023offsite}, and LN-TUNE~\cite{LNTUNE}. To evaluate the effectiveness of data protection, we extend the state-of-the-art data reconstruction attacks---FORA~\cite{fora}---to ViT. More details are provided in the Supplementary Materials.

\smallskip \noindent \textbf{Evaluation Metrics.} To evaluate the effectiveness of the target task adaptation, we use classification accuracy to measure the performance. As for the quality measurement of data reconstruction results, we adopt three metrics: SSIM~\cite{ssim} (S), PSNR~\cite{psnr} (P), and LPIPS~\cite{lpips} (L).

\begin{figure*}[h]
\centering
\begin{minipage}[h]{0.28\textwidth}
\centering
\captionof{table}{Comparison of client memory (MB) and computation costs (Min) between \SA{} and other baselines.}
\vspace{-8pt}
\hspace{-8pt}
\label{tab:computation_cost}
\resizebox{1.\textwidth}{!}{
\setlength{\tabcolsep}{.4mm}{
\begin{tabular}{l|c|c}
\toprule
Methods & GPU Memory & Training Time \\ \midrule
Linear Probing & $6979$ & $66$  \\
Fine Tuning & $10302$ &  $34$ \\
LN-TUNE & $5968$ & $25$  \\
Split Learning & $8932$ & $15$   \\
Offsite Tuning & $4896$  &$23$  \\ \midrule
\SA{} (ours) & $2233$ &  $2.5$\\ \bottomrule
\end{tabular}}}
\end{minipage}
\hspace{0.01\textwidth}
\begin{minipage}[h]{0.34\textwidth}
\centering
\vspace{-9.5pt}
\captionof{table}{Risk assessment of 3 most potential cases over 3 datasets (domains) for obtaining high-quality models from our \SA{} approach.}
\vspace{-2pt}
\hspace{-8pt}
\label{tab:model_protection}
\resizebox{1.\textwidth}{!}{
\setlength{\tabcolsep}{.5mm}{
\begin{tabular}{l|c|c|c}
\toprule
Variations & CIFAR-100 & Places365 & D.Net-Cl \\ \midrule
Quant. Frontend & $26.34$ & $13.45$ & $26.70$ \\
Original Frontend & $26.53$ & $13.72$ &$27.12$ \\
Auxiliary Backend & $30.65$ & $10.15$& $16.56$ \\ \midrule
\SA{} (ours) & $81.98$ & $35.31$ & $71.13$ \\ \bottomrule
\end{tabular}}}
\end{minipage}
\hspace{0.01\textwidth}
\begin{minipage}[h]{0.34\textwidth}
\centering
\captionof{table}{Ablation study of the major components of our \SA{} approach in 5-shot adaptation over 3 datasets (domains).}
\vspace{-2pt}
\hspace{-8pt}
\label{tab:ablation_study}
\resizebox{1.\textwidth}{!}{
\setlength{\tabcolsep}{.5mm}{
\begin{tabular}{l|c|c|c}
\toprule
Variations & CIFAR-100 & Places365 & D.Net-Cl \\ \midrule
\SA{} w/o HT Aug &  $80.37$&$35.27$  &$64.59$  \\
\SA{} w/o OOD QAT &  $79.03$&$29.50$ &$64.42$  \\
\SA{} w/o QAT &  $80.05$&$29.59$  & $64.16$  \\
\SA{} w/o PR Aug & $79.90$ & $29.56$  &  $52.85$ \\ \midrule
\SA{} (ours) & $81.98$& $35.31$& $71.13$\\ \bottomrule
\end{tabular}}}
\end{minipage}
\end{figure*}

\subsection{Effectiveness of \SA{} for Task Adaptation}
We create three few-shot adaptation scenarios for each client dataset or domain: 3-shot, 5-shot, and 10-shot per class, with experimental results shown in Table~\ref{tab:5-shot}. We can easily observe that our \SA{} can always substantially exceed all other baseline approaches. In particular, methods like Fine Tuning, LN-TUNE, and Offsite Tuning lag far behind \SA{}, which may be attributed to the overfitting caused by the low-level feature tuning. The method closest to \SA{} in terms of performance is Split Learning, but its biggest issue is the privacy leakage, which we will discuss in Section~\ref{sec:exp_privacy_protection}. In Table~\ref{tab:computation_cost}, we also provide the client memory and computation costs of each method (assuming the first three methods are conducted on the client side), from which we can observe that \SA{} even runs with the smallest cost. Note that \SA{} has no training on the client, thus we view the representation extraction time as the training time in Table~\ref{tab:computation_cost}.

\begin{table*}[h]
\centering
\caption{Comparison of defensive performance against the state-of-the-art data reconstruction attack---FORA~\cite{fora}---between \SA{} and Split Learning. We visualize the reconstruction and calculate SSIM (S$\downarrow$), PSNR (P$\downarrow$), and LPIPS (L$\uparrow$) to measure the reconstruction quality.}
\vspace{-8pt}
\hspace{-5pt}
\resizebox{1.\textwidth}{!}{
\setlength{\tabcolsep}{.5mm}{
\begin{tabular}{c|cc|cc|cc}
\toprule
\multicolumn{1}{c|}{Dataset} & \multicolumn{2}{c|}{CIFAR-100} & \multicolumn{2}{c|}{Places365} & \multicolumn{2}{c}{D.Net-Cl} \\ \midrule
\raisebox{-0.2\height}{\begin{tabular}[c]{@{}c@{}}Ground\\ Truth\end{tabular}} &\raisebox{-0.5\height}{\includegraphics[width=4cm]{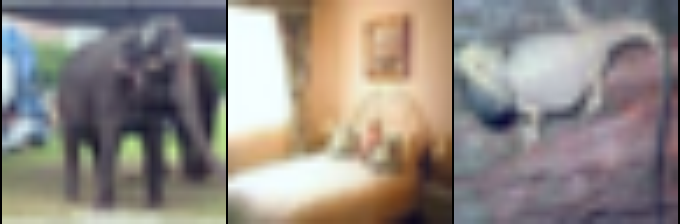}}  &\raisebox{-0.2\height}{-}  &\raisebox{-0.5\height}{\includegraphics[width=4cm]{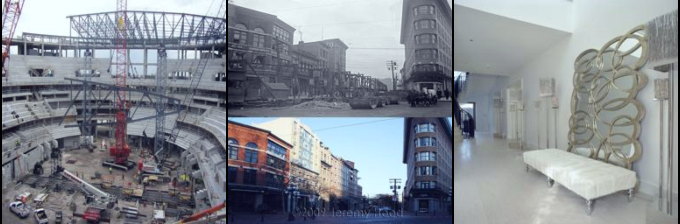}}  &\raisebox{-0.2\height}{-}  &\raisebox{-0.5\height}{\includegraphics[width=4cm]{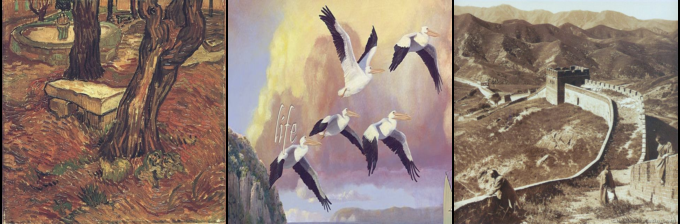}}  &\raisebox{-0.2\height}{-}  \\ \midrule
\raisebox{-0.2\height}{\begin{tabular}[c]{@{}c@{}}Split\\ Learning\end{tabular}} &\raisebox{-0.5\height}{\includegraphics[width=4cm]{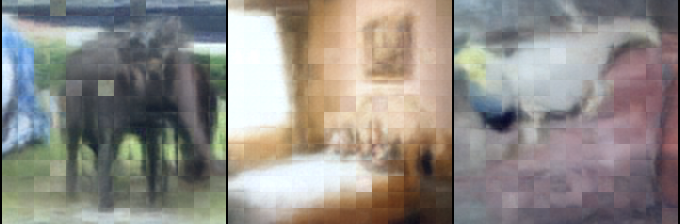}}  &\raisebox{-0.2\height}{\begin{tabular}[c]{@{}c@{}}S:$0.80$\\ P:$26.1$\\ L:$0.46$\end{tabular}}  &\raisebox{-0.5\height}{\includegraphics[width=4cm]{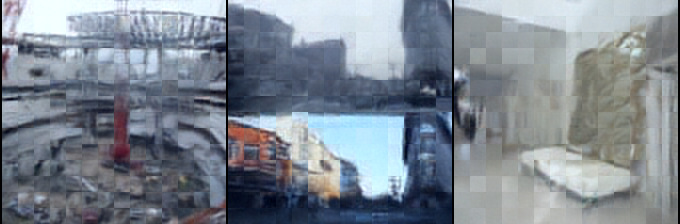}}  &\raisebox{-0.2\height}{\begin{tabular}[c]{@{}c@{}}S:$0.43$\\ P:$20.8$\\ L:$0.67$\end{tabular}}  &\raisebox{-0.5\height}{\includegraphics[width=4cm]{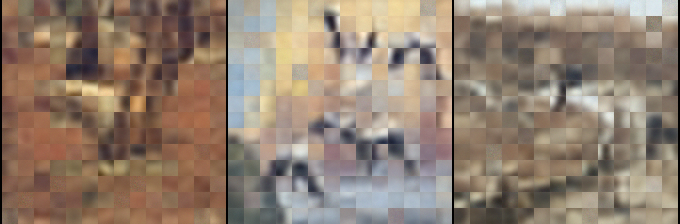}}  &\raisebox{-0.2\height}{\begin{tabular}[c]{@{}c@{}}S:$0.59$\\ P:$22.1$\\ L:$0.46$\end{tabular}}  \\ \midrule
\raisebox{-0.2\height}{\SA{} (ours)} &\raisebox{-0.5\height}{\includegraphics[width=4cm]{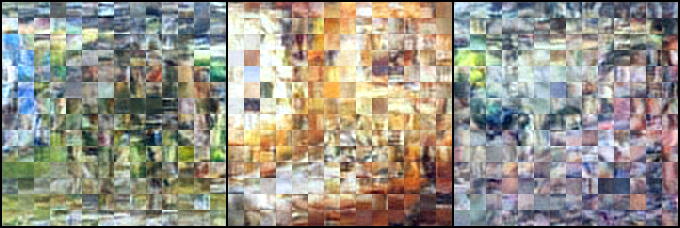}}  &\raisebox{-0.2\height}{\begin{tabular}[c]{@{}c@{}}S:$0.23$\\ P:$15.2$\\ L:$0.70$\end{tabular}}  &\raisebox{-0.5\height}{\includegraphics[width=4cm]{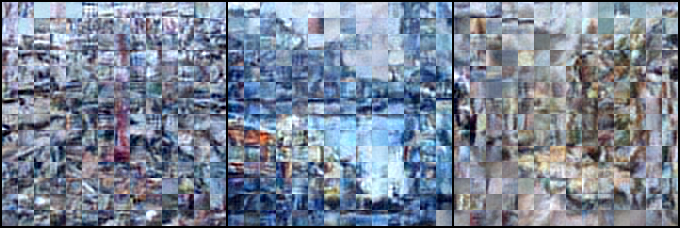}}  &\raisebox{-0.2\height}{\begin{tabular}[c]{@{}c@{}}S:$0.30$\\ P:$16.6$\\ L:$0.73$\end{tabular}}  &\raisebox{-0.5\height}{\includegraphics[width=4cm]{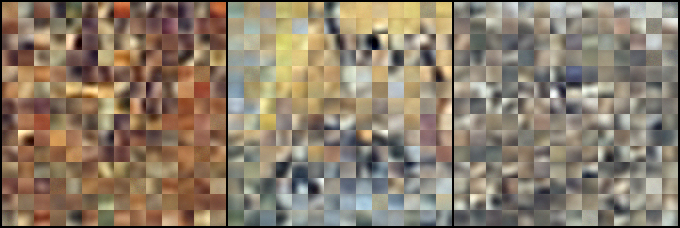}}  &\raisebox{-0.2\height}{\begin{tabular}[c]{@{}c@{}}S:$0.25$\\ P:$14.4$\\ L:$0.60$\end{tabular}}  \\ \bottomrule
\end{tabular}}}
\label{tab:data_reconstruction}
\vspace{-10pt}
\end{table*}
\begin{table}[h]
\centering
\caption{Ablation study of \SA{}'s bi-level noise when defending against FORA~\cite{fora} on CIFAR-100.}
\vspace{-8pt}
\hspace{-5pt}
\resizebox{.94\linewidth}{!}{
\setlength{\tabcolsep}{.5mm}{
\begin{tabular}{c|cc}
\toprule
\multicolumn{1}{c|}{Dataset} & \multicolumn{2}{c}{CIFAR-100} \\ \midrule
\raisebox{-0.2\height}{\begin{tabular}[c]{@{}c@{}}Ground\\ Truth\end{tabular}} &\raisebox{-0.5\height}{\includegraphics[width=4cm]{figures/cifar-gt.png}}  &\raisebox{-0.2\height}{-}   \\ \midrule
\raisebox{-0.2\height}{\begin{tabular}[c]{@{}c@{}}\SA{} w/o\\ Laplace Noise\end{tabular}} &\raisebox{-0.5\height}{\includegraphics[width=4cm]{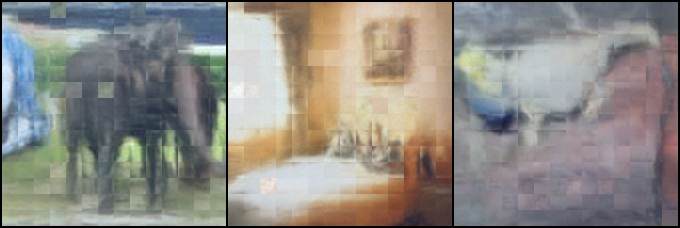}}  &\raisebox{-0.2\height}{\begin{tabular}[c]{@{}c@{}}S:$0.76$\\ P:$23.1$\\ L:$0.49$\end{tabular}}  \\ \midrule
\raisebox{-0.2\height}{\begin{tabular}[c]{@{}c@{}}\SA{} w/o\\ Model Noise\end{tabular}} &\raisebox{-0.5\height}{\includegraphics[width=4cm]{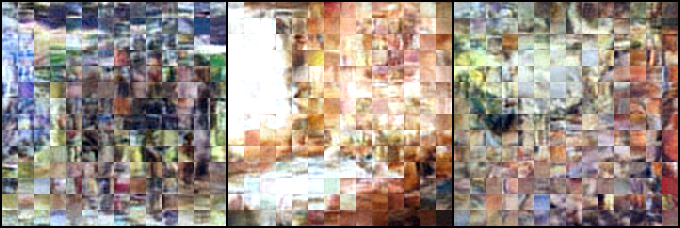}}  &\raisebox{-0.2\height}{\begin{tabular}[c]{@{}c@{}}S:$0.28$\\ P:$16.4$\\ L:$0.67$\end{tabular}}  \\ \midrule
\raisebox{-0.2\height}{\SA{} (ours)} &\raisebox{-0.5\height}{\includegraphics[width=4cm]{figures/cifar-noise.png}}  &\raisebox{-0.2\height}{\begin{tabular}[c]{@{}c@{}}S:$0.23$\\ P:$15.2$\\ L:$0.70$\end{tabular}}  \\ \bottomrule
\end{tabular}}}
\label{tab:data_reconstruction_ablation}
\vspace{-10pt}
\end{table}

\subsection{Effectiveness of \SA{} for Server Pre-trained ViT and Client Data Protection}
\label{sec:exp_privacy_protection}
For privacy protection, we first test whether the quantized frontend of \SA{} can result in high-quality models with good downstream task performance. We consider the three most potential variations: 1) directly tune a task module on the quantized frontend (Quant. Frontend); 2) tune a task module on the original frontend though this case is impossible in practice (Original Frontend); 3) attach a backend from other models and tune a task module on the quantized frontend and auxiliary backend (Auxiliary Backend). The experimental results are shown in Table~\ref{tab:model_protection}, in which all variations perform much poorer than Linear Probing in Table~\ref{tab:5-shot} and our \SA{}. In this case, we believe \SA{} can protect the pre-trained model to a great extent.

Then we launch FORA~\cite{fora} to evaluate the data protection of our \SA{}. For a fair comparison, only Split Learning is compared with \SA{} as only it requires the client to share intermediate data representations with the server. The experiment results are presented in Table~\ref{tab:data_reconstruction}. First, we can visually observe that the reconstruction results of \SA{} are much worse than those of Split Learning. Then the three metrics also demonstrate the same phenomenon, i.e., FORA can only recover much lower-quality images when attacking \SA{} compared with Split Learning. In summary, our \SA{} offers strong protection of the client data. More experiments on the noise degree can be found in Supplementary Materials.

\subsection{Ablation Study}
We conduct comprehensive ablation studies in the 5-shot adaptation on CIFAR-100, Places365, and DomainNet-Cl.

\smallskip \noindent \textbf{Hilbert Transform-based Data Augmentation.} We detach the use of HT data augmentation in the frontend quantization and backend tuning (`\SA{} w/o HT Aug'). Compared with the full \SA{} in Table~\ref{tab:ablation_study}, the variation without HT data augmentation performs worse, which proves its gain in improving adaptation performance.

\smallskip \noindent \textbf{Out-of-distribution Quantization-aware Tuning.} We design two variations: one is not to use OOD quantization-aware tuning (`\SA{} w/o OOD QAT'), and the other is only to use the quantized frontend $\widehat{F}_\mathrm{F}$ characterized on the full server dataset (`\SA{} w/o QAT'). Compared with the full \SA{}, the effectiveness of the representation-mixup quantization-aware tuning and the simulated OOD frontends in alleviating quantization-induced performance drop can be verified.

\smallskip \noindent \textbf{Patch Retrieval Augmentation.} We don't use the patch retrieval augmentation and observe how the model performance will change. The big difference between `\SA{} w/o PR Aug' and `\SA{}' in Table~\ref{tab:ablation_study} indicates the essential role of our augmentation here in performance enhancement.

\smallskip \noindent \textbf{Bi-level Noisy Representation Extraction.} We detach the use of model Gaussian noise (`\SA{} w/o Model Noise') and representation Laplace noise (`\SA{} w/o Laplace Noise'), respectively, and the experiments are shown in Table~\ref{tab:data_reconstruction_ablation}. Model noise and representation noise play a synergistic role greater than the simple sum of them in data protection.
\section{Conclusion}
In this work, we propose Split Adaptation (\SA{}) to adapt pre-trained ViTs for downstream tasks while protecting the data and model in the sever-client framework. After dividing the ViT into a frontend and a backend, \SA{} innovatively uses model quantization to protect the frontend. \SA{} also allows the client to introduce a bi-level noise to the quantized frontend and the extracted data representations for data protection. Accordingly, \SA{} uses data-level and model-level augmentation to mitigate the influence of the quantization and the noise. Besides, \SA{} proposes patch retrieval augmentation to address overfitting in few-shot adaptation. Extensive experiments have validated the superiority of \SA{} in task adaptation and data-model protection.

{
    \small
    \bibliographystyle{ieeenat_fullname}
    \bibliography{main}
}

\clearpage
\setcounter{page}{1}


\appendix
\onecolumn
\section*{\Large \centering{Appendix}}


\section*{Overview}
This appendix contains additional details of the paper `\textit{Split Adaptation for Pre-trained Vision Transformers}', including more implementation details of the main experiments, additional experiment results, and sensitivity analysis of parameters used in our methods.
\begin{itemize}
    \item Section~\ref{sec:app_implementation} provides more implementation details and setups of our \SA{} (Section~\ref{sec:app_setup_SA}) and comparison state-of-the-art baseline approaches (Section~\ref{sec:app_baselines}).
    \item Section~\ref{sec:app_experiments} reports additional experiment results of performing adaptation on more ViT architectures: Deit-Base and Swin-L.
    \item Section~\ref{sec:app_sensitivity} shows the sensitivity analysis on the bi-level noise degree (Section~\ref{sec:app_noise_degree}) and patch retrieval augmentation (Section~\ref{sec:app_patch_aug}) in our \SA{}.
\end{itemize}

\section{More Implementation Details}
\label{sec:app_implementation}
\subsection{More setups of \SA{}}
\label{sec:app_setup_SA}
In addition to the setups mentioned in the main paper, we provide more setups of \SA{} here. In the OOD quantization-aware tuning, we cut the merged server dataset into $M=3$ subsets and quantize $M=3$ frontends accordingly. As for the random seeds, we randomly use 1, 42, 215 to run experiments repeatedly. Moreover, in the risk assessment of the model protection provided by \SA{}, we adopt the last 1/3 layers of DinoV2-Large~\cite{dino} as the auxiliary backend.

\subsection{Implementation of Baseline Approaches}
\label{sec:app_baselines}
All baseline approaches along with our \SA{} adopt the same task module, i.e., two linear layers for the Places365 and a single linear layer for other datasets. The batch size we used is 32 and we set the training epochs as when the training loss does not decrease. We follow the default setups of the baseline approaches when setting the optimizer and its learning rate. 

\smallskip \noindent \textbf{Linear Probing~\cite{linear_probing}:} Following the standard setups, we freeze the pre-trained ViT and only tune the task module in 100 epochs.

\smallskip \noindent \textbf{Fine Tuning:} After attaching the task module behind the pre-trained ViT, we forward the client data to the model and tune it entirely in 100 epochs.

\smallskip \noindent \textbf{LN-TUNE~\cite{LNTUNE}:} LN-TUNE only tunes the LayerNorm parameters in transformers, thus we filter out these parameters and tune them along with the task module in 100 epochs.

\smallskip \noindent \textbf{Split Learning~\cite{vepakomma2018split}:} The same splitting location as our \SA{} is adopted here, i.e., cut the first 2/3 layers of the pre-trained ViT as the frontend, while the rest 1/3 layers form the backend. The client manages the frontend and the task module, while the server hosts the backend. Following the default settings, the backend is frozen, and only the frontend and the task module are optimized by back-propagating the training loss in 100 epochs.

\smallskip \noindent \textbf{Offsite Tuning~\cite{xiao2023offsite}:} Following the default setups, we adopt the layer-drop strategy to implement Offsite Tuning. Specifically, we regard the first and last two layers of the pre-trained ViT as the adapters that the client will tune. Then the emulator is formed by randomly dropping half intermediate layers (except the layers selected as the adapters). The adapter along with the task module is tuned in 50 epochs on the client side.

\smallskip \noindent \textbf{Data Reconstruction Attack---FORA~\cite{fora}:} Following the training workflow in FORA~\cite{fora}, we use the frontend model from the server side as the substitute model. In FORA, Multi-Kernel Maximum Mean Discrepancy
(MK-MMD) and discriminator losses are used to align the attacker encoder with the client encoder. Since the client does not share its encoder with the server, we designed to use the server frontend directly as the attacker encoder. During training, the frontend is frozen, thus these losses are no longer necessary. We remove the MK-MMD and discriminator during training. We developed an inverse model using decoder layers similar to the pre-trained ViT encoder layers, which are trained to reconstruct data from the bi-level noisy representations.

\section{Additional Experiments}
\label{sec:app_experiments}
\begin{table*}[h]
\centering
\caption{Performance comparison between our \SA{} and other baseline approaches in 3-shot, 5-shot, and 10-shot adaptation scenarios on Deit-Base. The \SA{} can substantially exceed other methods. We bold and blue \blue{\textbf{the best}}, and bold \textbf{the second best}.}
\vspace{-8pt}
\hspace{-8pt}
\resizebox{.95\textwidth}{!}{
\setlength{\tabcolsep}{1.2mm}{
\centering
\begin{tabular}{l|ccc|ccc|ccc}
\toprule
Few-shot Setup & \multicolumn{3}{c|}{3-shot} & \multicolumn{3}{c|}{5-shot} & \multicolumn{3}{c}{10-shot}\\ \midrule
Method & CIFAR-100 & Places365 & D.Net-Cl & CIFAR-100 & Places365 & D.Net-Cl & CIFAR-100 & Places365 & D.Net-Cl \\ \midrule
Linear Probing &\blue{$\bm{50.58}_{\pm 1.93}$}&$\bm{19.13}_{\pm 0.53}$&\blue{$\bm{34.65}_{\pm 1.67}$}&$\bm{56.80}_{\pm 0.87}$&$\bm{22.22}_{\pm 0.12}$&$\bm{42.19}_{\pm 0.28}$&$\bm{63.24}_{\pm 0.80}$&$26.99_{\pm 0.11}$&$50.31_{\pm 1.07}$    \\

Fine Tuning &$24.45_{\pm 2.91}$&$15.28_{\pm 0.10}$&$16.25_{\pm 0.83}$&$36.24_{\pm 2.31}$&$20.60_{\pm 0.75}$&$32.36_{\pm 1.42}$&$57.54_{\pm 1.62}$&$\bm{28.05}_{\pm 0.70}$&$52.38_{\pm 1.66}$    \\

LN-TUNE &$11.17_{\pm 1.71}$&$0.96_{\pm 0.28}$&$2.98_{\pm 0.23}$&$16.20_{\pm 2.02}$&$1.55_{\pm 0.45}$&$4.72_{\pm 0.66}$&$18.82_{\pm 5.58}$&$1.86_{\pm 1.40}$& $7.34_{\pm 0.96}$   \\

Split Learning &$20.55_{\pm 3.68}$&$16.32_{\pm 0.82}$&$20.55_{\pm 0.78}$&$30.37_{\pm 2.07}$&$20.79_{\pm 0.96}$&$\bm{42.19}_{\pm 0.28}$&$42.94_{\pm 0.58}$&$26.14_{\pm 1.05}$&$48.33_{\pm 3.03}$    \\

Offsite Tuning &$26.18_{\pm1.99}$&$15.64_{\pm1.67}$&$21.40_{\pm1.16}$&$37.20_{\pm3.06}$&$21.19_{\pm0.41}$&$33.42_{\pm2.20}$&$58.14_{\pm2.40}$&$27.84_{\pm0.60}$&$\bm{52.58}_{\pm0.74}$    \\ \midrule

\SA{} (ours) &$\bm{48.86}_{\pm 1.60}$&\blue{$\bm{21.70}_{\pm 0.52}$}&$\bm{34.55}_{\pm 1.49}$&\blue{$\bm{58.22}_{\pm 0.79}$}&\blue{$\bm{26.16}_{\pm 0.16}$}&\blue{$\bm{46.64}_{\pm 0.34}$}&\blue{$\bm{68.90}_{\pm 0.57}$}&\blue{$\bm{31.75}_{\pm 0.31}$}&\blue{$\bm{58.27}_{\pm 0.49}$}    \\ 

\bottomrule
\end{tabular}}}
\label{tab:deit-b}
\end{table*}
\begin{table*}[h]
\centering
\caption{Performance comparison between our \SA{} and other baseline approaches in 3-shot, 5-shot, and 10-shot adaptation scenarios on Swin-L. The \SA{} can substantially exceed other methods. We bold and blue \blue{\textbf{the best}}, and bold \textbf{the second best}.}
\vspace{-8pt}
\hspace{-8pt}
\resizebox{.95\textwidth}{!}{
\setlength{\tabcolsep}{1.2mm}{
\centering
\begin{tabular}{l|ccc|ccc|ccc}
\toprule
Few-shot Setup & \multicolumn{3}{c|}{3-shot} & \multicolumn{3}{c|}{5-shot} & \multicolumn{3}{c}{10-shot}\\ \midrule
Method & CIFAR-100 & Places365 & D.Net-Cl & CIFAR-100 & Places365 & D.Net-Cl & CIFAR-100 & Places365 & D.Net-Cl \\ \midrule
Linear Probing &\blue{$\bm{71.28}_{\pm1.79}$}&\blue{$\bm{33.17}_{\pm0.54}$}&$\bm{58.22}_{\pm0.80}$&$\bm{75.76}_{\pm1.07}$&$\bm{36.95}_{\pm0.36}$&$64.76_{\pm0.79}$&$79.83_{\pm0.40}$&$\bm{40.62}_{\pm0.25}$&  $70.40_{\pm 0.26}$  \\
Fine Tuning &$61.23_{\pm2.53}$&$29.12_{\pm0.42}$&$55.74_{\pm0.64}$&$73.05_{\pm1.23}$&$35.09_{\pm0.32}$&$\bm{65.26}_{\pm1.63}$&\blue{$\bm{82.65}_{\pm0.41}$}&$39.79_{\pm0.17}$&\blue{$\bm{72.92}_{\pm1.06}$}    \\
LN-TUNE &$28.98_{\pm7.47}$&$4.32_{\pm1.62}$&$7.32_{\pm3.20}$&$29.23_{\pm16.2}$&$6.35_{\pm1.64}$&$12.45_{\pm3.63}$&$27.42_{\pm11.8}$&$8.12_{\pm2.24}$&$17.72_{\pm11.0}$    \\
Split Learning &$51.51_{\pm4.66}$&$25.76_{\pm0.22}$&$49.64_{\pm2.66}$&$65.37_{\pm1.41}$&$32.19_{\pm0.43}$&$59.36_{\pm2.73}$&$73.59_{\pm2.57}$&$38.30_{\pm0.44}$&$70.33_{\pm1.38}$    \\
Offsite Tuning &$62.23_{\pm2.27}$&$29.66_{\pm0.51}$&$55.03_{\pm0.87}$&$72.80_{\pm0.65}$&$35.75_{\pm0.49}$&$65.20_{\pm0.90}$&$80.68_{\pm0.83}$&$39.71_{\pm0.83}$&$71.36_{\pm1.05}$    \\ \midrule
\SA{} (ours) &$\bm{70.22}_{\pm0.92}$&$\bm{31.87}_{\pm0.31}$&\blue{$\bm{59.83}_{\pm1.61}$}&\blue{$\bm{76.66}_{\pm0.75}$}&\blue{$\bm{37.20}_{\pm0.28}$}&\blue{$\bm{66.13}_{\pm0.92}$}&$\bm{80.96}_{\pm0.28}$&\blue{$\bm{41.00}_{\pm0.44}$}&$\bm{72.13}_{\pm0.59}$    \\ \bottomrule
\end{tabular}}}
\label{tab:swin_l}
\vspace{-5pt}
\end{table*}
\begin{figure}[t]
    \centering
    \begin{subfigure}[b]{0.33\linewidth}
        \includegraphics[width=\linewidth]{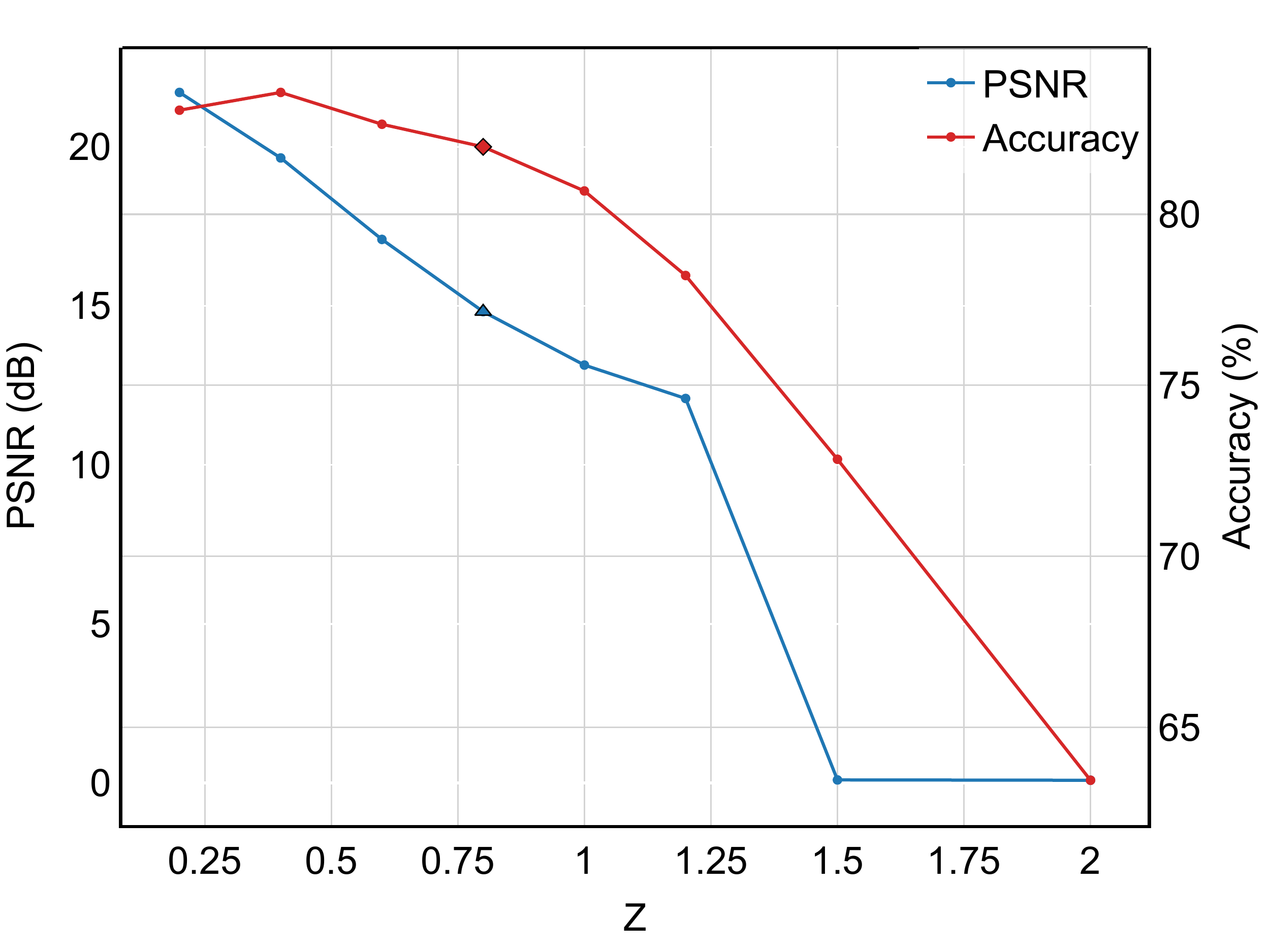}
        \vspace{-15pt}
        \caption{PSNR}
    \end{subfigure}
    \hfill
    \begin{subfigure}[b]{0.33\linewidth}
        \includegraphics[width=\linewidth]{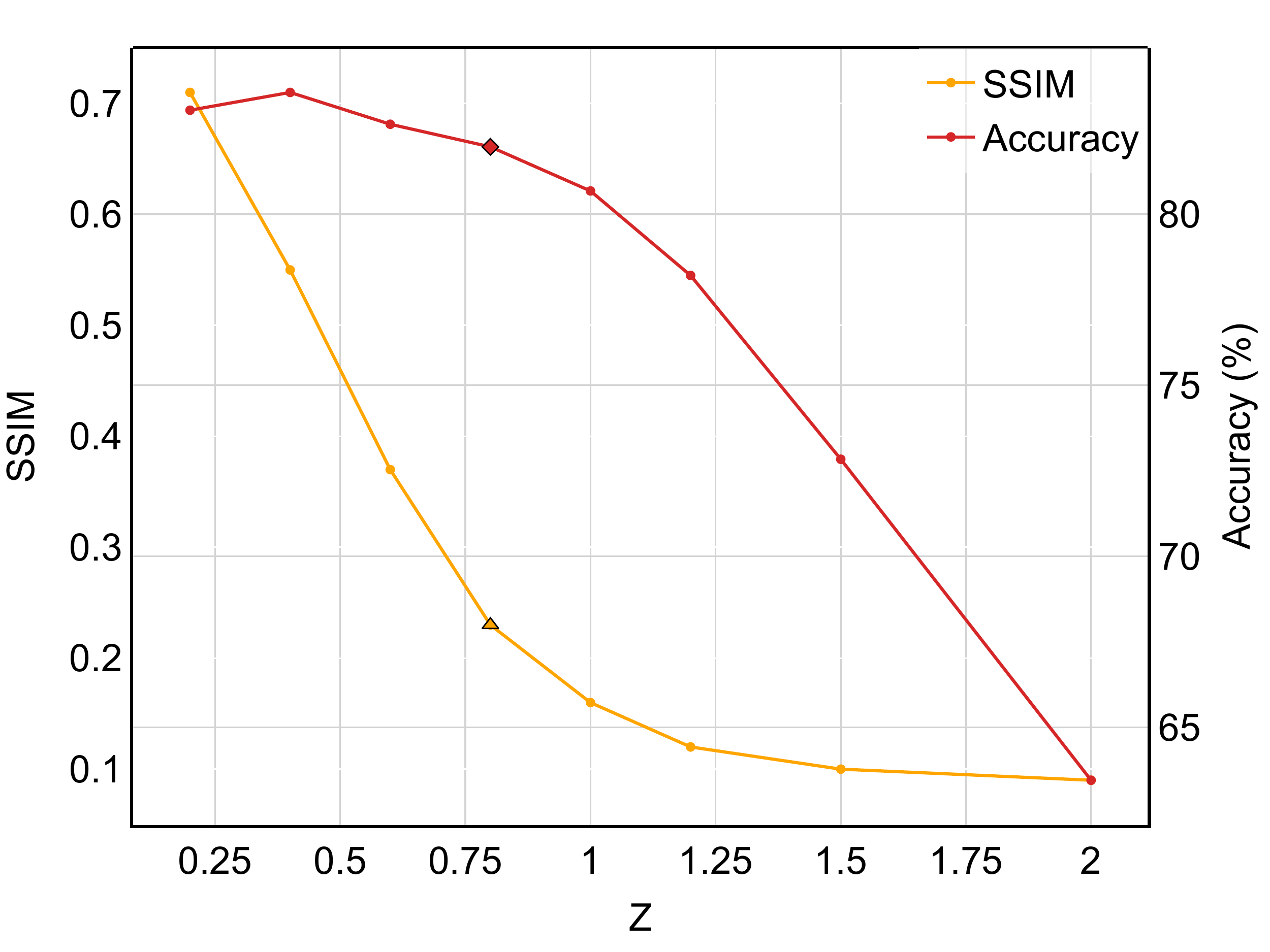}
        \vspace{-15pt}
        \caption{SSIM}
    \end{subfigure}
    \hfill
    \begin{subfigure}[b]{0.33\linewidth}
        \includegraphics[width=\linewidth]{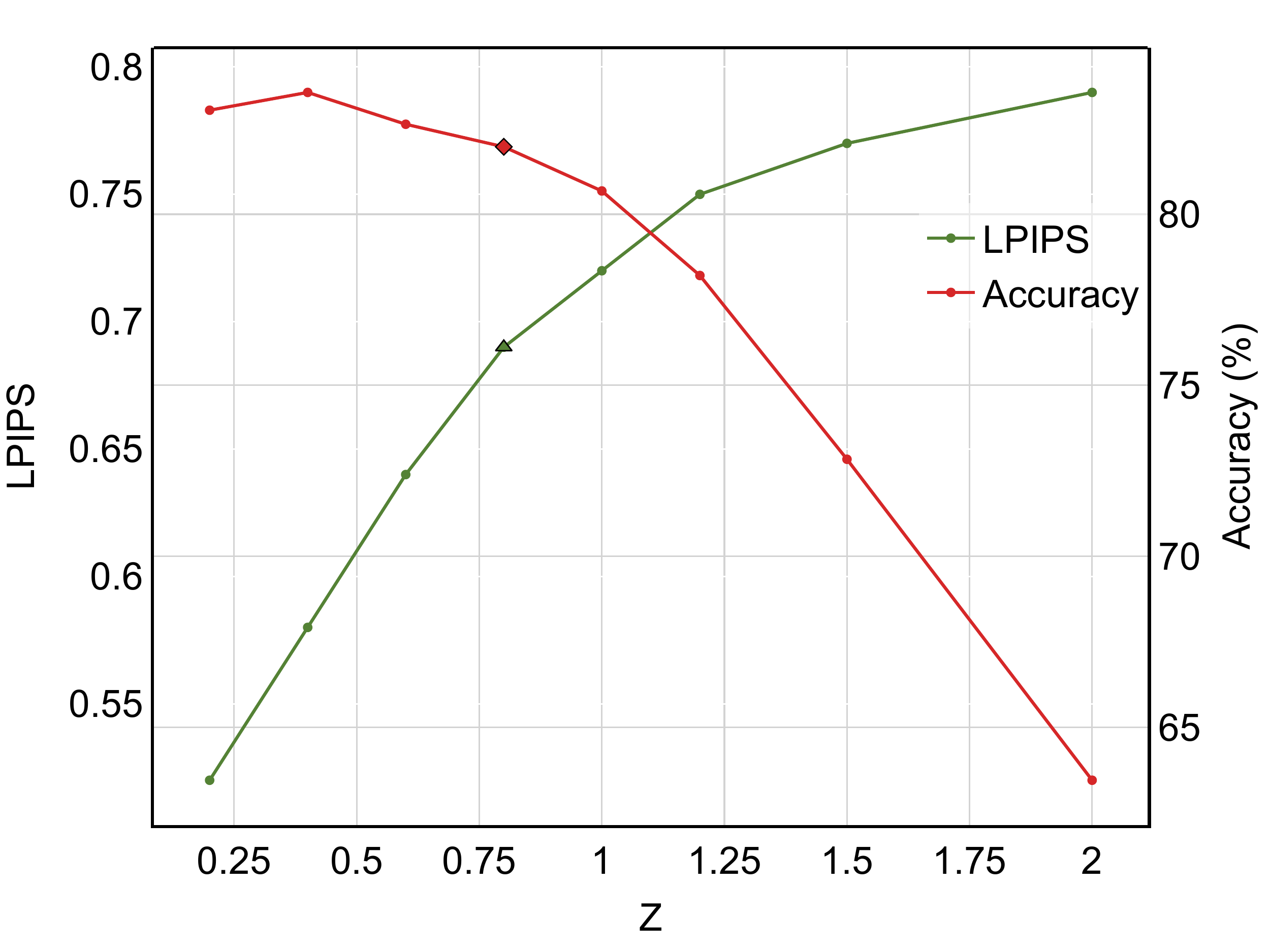}
        \vspace{-15pt}
        \caption{LPIPS}
    \end{subfigure}
    \vspace{-20pt}
    \caption{Sensitivity analysis of bi-level noisy representation extraction by changing the injected Laplace noise degree with a variety of z.}
    \label{fig:noise_sensitivity}
\end{figure}
\begin{table*}[htbp]
\centering
\caption{Sensitivity analysis of patch number $N^\mathrm{P}$ in our patch retrieval augmentation when setting the augmentation runs as $N^\mathrm{Aug}=64$.}
\vspace{-8pt}
\hspace{-8pt}
\resizebox{.95\textwidth}{!}{
\setlength{\tabcolsep}{1.5mm}{
\centering
\begin{tabular}{c|ccccccccccccccc}
\toprule
$N^\mathrm{P}$ & 0 & 10 & 20 & 30 & 40 & 50 & 60 & 70 & 80 & 90 & 100 & 120 & 140 & 160 & 180 \\ \midrule
\SA{} &$80.41$  &$81.60$  &$80.61$  &$80.88$  &$80.61$  &$82.52$  &$80.11$  &$82.33$  &$80.78$  &$82.18$  &$81.54$  &$81.86$  &$81.84$  &$81.92$  &$81.52$  \\ \bottomrule
\end{tabular}}}
\label{tab:app_sen_N_P}
\end{table*}
\begin{table*}[htbp]
\centering
\caption{Sensitivity analysis of augmentation runs $N^\mathrm{Aug}$ in our patch retrieval augmentation when setting the patch number as $N^\mathrm{P}=50$.}
\vspace{-8pt}
\hspace{-8pt}
\resizebox{.95\textwidth}{!}{
\setlength{\tabcolsep}{1.5mm}{
\centering
\begin{tabular}{c|ccccccccccccccc}
\toprule
$N^\mathrm{Aug}$ & 0 & 1 & 2 & 4 & 8 & 20 & 30 & 40 & 50 & 60 & 64 & 70 & 80 & 90 & 100 \\ \midrule
\SA{} &$80.41$  &$79.94$  &$81.02$  &$81.83$  &$81.95$  &$81.41$  &$81.24$  &$80.83$  &$80.85$  &$81.15$  &$82.52$  &$81.14$  &$81.49$  &$82.26$  &$80.84$  \\ \bottomrule
\end{tabular}}}
\label{tab:app_sen_N_aug}
\end{table*}

In this section, we experimented with more model architectures: Deit-Base~\cite{deit} and Swin-L~\cite{liu2021swin}. For Deit-Base, we also split the first 2/3 layers as the frontend and view the rest 1/3 layers as the backend. As for Swin-L, it consists of four stages, and each stage forms a sub-transformer with unique setups of embedding dimensions and attention heads. Therefore, we cut the Swin-L stage-wise, i.e., splitting the first three stages as the frontend while the last stage is the backend. Besides, due to the window sliding mechanism in the swin-transformer, the patch sequence of each sample is uncertain, making our patch retrieval augmentation inapplicable. The setups of the task module are the same as those of ViT-L. The adaptation scenarios are still 3-shot, 5-shot, and 10-shot. According to the experimental results shown in Tables~\ref{tab:deit-b} and \ref{tab:swin_l}, we can clearly observe that our \SA{} approach achieves the best performance in almost all cases, in particular, note that \SA{} does not apply patch retrieval augmentation in experiments of Swin-L. Methods whose tuning spreads over the entire model, like Fine Tuning, LN-TUNE, and Offsite Tuning, perform much poorer than others. The potential reason is still the overfitting of the limited client data. As for the partial tuning methods, like Linear Probing and Split Learning, they are influenced much less by overfitting but still lag far behind our \SA{}.

\section{Sensitivity Analysis}
\label{sec:app_sensitivity}
\subsection{Analysis of Bi-level Noise Degree}
\label{sec:app_noise_degree}
In \SA{}, the bi-level noise (Section~\ref{sec:client_side}) directly impacts the adaptation performance and the defensive capability regarding data reconstruction attacks. As a result, we conduct a sensitivity analysis on CIFAR-100 by changing the noise degree. Specifically, we adopt a series of Laplace noises---$\mathrm{Laplace}(0, 0.2)$, $\mathrm{Laplace}(0, 0.4)$, $\mathrm{Laplace}(0, 0.6)$, $\mathrm{Laplace}(0, 0.8)$, $\mathrm{Laplace}(0, 1.0)$, $\mathrm{Laplace}(0, 1.2)$, $\mathrm{Laplace}(0, 1.5)$, and $\mathrm{Laplace}(0, 2.0)$---to conduct the pre-trained ViT adaptation and then launch FORA to reconstruct the client data. We calculate the three metrics to measure the reconstruction quality and associate them with the adaptation performance to draw the trade-off curves in Figure~\ref{fig:noise_sensitivity}. According to these experimental results, the noise $\mathrm{Laplace}(0, 0.8)$ achieves a good trade-off between adaptation performance and client data protection.

\subsection{Analysis of Patch Retrieval Augmentation}
\label{sec:app_patch_aug}
The patch retrieval augmentation (Section~\ref{sec:patch_aug}) in \SA{} requires two hyperparameters: one is the number of the patches $N^\mathrm{P}$ that need retrieval and replacement, the other is the number of augmentation runs $N^\mathrm{Aug}$. We conduct a sensitivity analysis of $N^\mathrm{P}$ and $N^\mathrm{Aug}$ on CIFAR-100. We first empirically set $N^\mathrm{Aug}=64$ and adopt a series of $N^\mathrm{P}$s. Then we select a patch number $N^\mathrm{P}=50$ that performs relatively well and change the $N^\mathrm{Aug}$ to conduct experiments. The results are shown in Tables~\ref{tab:app_sen_N_P} and \ref{tab:app_sen_N_aug}, in which we can observe that no matter what setups of $N^\mathrm{P}$ and $N^\mathrm{Aug}$, our patch retrieval augmentation is generally effective in enhancing the adaptation though there is certain fluctuation. Then we adopt an empirically good setup ($N^\mathrm{P}=50$ and $N^\mathrm{Aug}=64$) in our main experiments.


\end{document}